\documentclass{svproc}
\usepackage{cite}
\usepackage{amsmath,amssymb,amsfonts}
\usepackage{algorithmic}
\usepackage{graphicx}
\usepackage{textcomp}
\usepackage{xcolor}
\usepackage{comment}
\usepackage{pdfpages}

\begin{document}

\mainmatter              
\title{Integrating White and Black Box Techniques for Interpretable Machine Learning}
\titlerunning{Integrating White and Black Box Techniques for Interpretable ML}  
%
\author{Eric M. Vernon \and Naoki Masuyama \and Yusuke Nojima}
\authorrunning{E. M. Vernon et al.} 
%
\tocauthor{Eric M. Vernon, Naoki Masuyama, and Yusuke Nojima}
\institute{Osaka Metropolitan University, Sakai, Osaka 5998531, Japan,\\
\email{\{sn22864k@st., masuyama@, nojima@\}omu.ac.jp}\\ 
}

\maketitle              

\begin{abstract}
In machine learning algorithm design, there exists a trade-off between the interpretability and performance of the algorithm. In general, algorithms which are simpler and easier for humans to comprehend tend to show worse performance than more complex, less transparent algorithms. For example, a random forest classifier is likely to be more accurate than a simple decision tree, but at the expense of interpretability. In this paper, we present an ensemble classifier design which classifies easier inputs using a highly-interpretable classifier (i.e., white box model), and more difficult inputs using a more powerful, but less interpretable classifier (i.e., black box model).
\end{abstract}

\section{Introduction}
One of the most pressing issues in machine learning (ML) research today is the concern that many popular ML algorithms operate as a ``black box'' - that is, they offer no human-understandable explanation for their outputs. Generally speaking, there is a trade-off (the so-called ``accuracy-interpretability trade-off'') between the performance of an ML model and how easily a human can understand the steps taken to reach a given conclusion.

For example, this trade-off can be seen quite plainly in decision trees \cite{Bohanec1994}. A shallow decision tree with only a few branches is quite easy to understand, but is more limited in its ability to describe complex datasets. Increasing the depth of the tree will generally improve accuracy, but at the cost of interpretability. Ensemble methods such as random forests \cite{Breiman2001} or gradient boosted trees \cite{Friedman2001} are generally even more accurate while further obscuring the decision making. 

In this paper, we present an ensemble classifier design which uses a simple, easily understood classifier to classify ``easy'' inputs and a more complex classifier for ``hard'' inputs. The final piece of the ensemble is a ``grader'' classifier which classifies inputs as either ``easy'' or ``hard''.

To classify a new input using our design, it is first evaluated by the grader. If the output is ``easy'', then the pattern is evaluated by the ``base classifier'' (e.g., a decision tree). If the output is ``hard'', then the pattern is evaluated by the ``deferral classifier'' (e.g., a random forest). In our experiments, we use a decision tree classifier for the grader as well. This means that the user will either be able to directly understand the reasoning behind system output (in the case of easy inputs), or be given an understandable reason for why a more complex classifier was required (in the case of hard inputs).

To train the ensemble, the base and deferral classifiers are independently fit to the training data, as per normal. Then, the training data copied and assigned new labels, either ``easy'' or ``hard'': Easy inputs are patterns which are correctly classified by the base classifier, hard inputs are those which are not. The grader is then fit to this modified training set.

This paper is organized as follows: In Section 2, we describe the background of our research. Section 3 gives a detailed overview of our proposed method, including a worked example using a synthetic 2-D dataset. Section 4 describes the computational experiments performed to validate our approach and the associated results; Section 5 concludes the paper.

\section{Research Background}
\subsection{Interpretability in Machine Learning}
Machine learning algorithms are everywhere: They personalize the content we see on the web, secure our bank accounts against fraud, and assist our doctors in diagnosing our health conditions. Advances in techniques such as deep learning and gradient boosting have given rise to incredibly powerful algorithms in tasks such as classification and regression \cite{Bentejac2021}, reinforcement learning \cite{Silver2018}, and language generation \cite{Brown2020}. This, in turn, has led to the influx of algorithms in our daily lives.

Unfortunately, many of these powerful algorithms effectively operate as a black-box, offering little to no insight into the ``reasoning'' behind the output. The lack of transparency has both ethical and legal ramifications, while simultaneously reducing acceptance of algorithms' output, especially in fields such as finance and medicine \cite{Lin2019, Longoni2019, Shin2021}. These concerns have given rise to the study of ``explainable artificial intelligence''. In particular, the research community has given tremendous effort to the pursuit of explaining deep neural networks \cite{Wocjciech2021}. While there have been promising advances in this area, there is still a compelling argument for the use of algorithms which are interpretable by nature when the use case demands.

\subsection{Classification with a Reject Option}
One area of research in the classification domain is the reject option. This allows for the classifier to ``reject'' classification of a given input, effectively saying ``I don't know'' instead of outputting a low-probability guess \cite{Hendrickx2021}. The reject option is therefore a consideration in situations where the cost of a misclassification sufficiently outweighs the cost of rejecting classification (and for example, manually labeling the input).

Traditionally, the reject option is implemented via numerical thresholding. This is a very natural approach, especially when using probabalistic classifiers: With fixed costs for misclassification and rejection, the optimal reject threshold can be found mathematically \cite{Chow1970}. Even when using non-probabalistic classifiers, the reject option is often designed to target patterns when lie near the classifier's decision boundary \cite{Ishibuchi1998, Nojima2016}.

One drawback of this approach is that there is little explanation offered for why classification was rejected, beyond ``the input is near the decision boundary''. Using a second, interpretable classifier to decide whether classification should be attempted or rejected has the potential to offer the user a more meaningful explanation \cite{Nojima2023}.

Our proposed method draws inspiration from these ideas. Instead of choosing to either accept or reject classification, for any given input our method instead chooses whether a ``white box'' (i.e., interpretable-by-nature) or a ``black box'' (i.e., non-interpretable, but likely more powerful) model is most appropriate. Using a second white box model to make this determination allows for some transparency even when the black box model was selected.

\section{Proposed Method}
\subsection{Overview}
Our method creates an ensemble of three components:

\begin{itemize}
  \item The \textit{base classifier} (white box): This is the classifier responsible for assigning labels to ``easy'' patterns. The specific choice of classification algorithm and associated parameters is specified by the user, but it should be considered ``highly interpretable'' within the context of the problem. In our experiments, we use decision trees with a maximum of four binary splits (resulting in at most $2^5 - 1$ total nodes).
  
  \item The \textit{deferral classifier} (black box): This is the classifier responsible for assigning labels to ``hard'' patterns. Similar to the base classifier, the choice of algorithms is entirely user-specified. However, a high-performance classifier should be chosen, without considering the interpretability of its outputs. In our experiments, we use random forest classifiers.
  
  \item The \textit{grader} (white box): This is the classifier which is responsible for deciding if any given input is ``easy'' or ``hard''. As with the base and deferral classifiers, the algorithm and parameters are user-specified. In our experiments, we use decision trees with a maximum of four binary splits.
\end{itemize}

\begin{figure}[t]
\includegraphics[width=0.6\linewidth]{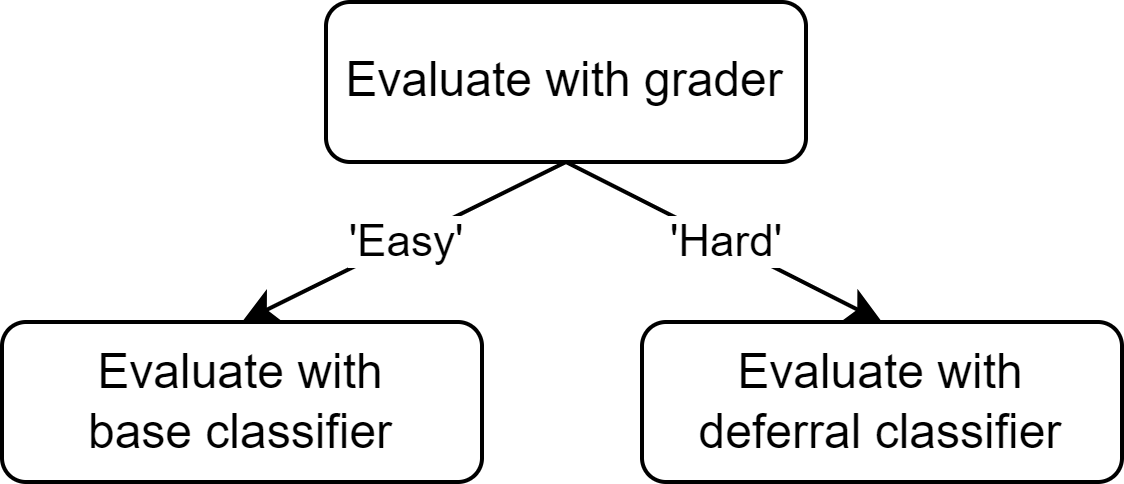}
\centering
\caption{The two-step process of evaluating new inputs.}
\label{fig:evaluation}
\end{figure}

\subsection{Training}
Training the ensemble is performed as follows:

\begin{enumerate}
  \item Initialize the base and deferral classifiers,  fit to the training data as usual.
  \item Relabel the training patterns as follows: Patterns correctly classified by the trained base classifier are ``easy''; patterns misclassified are ``hard''.
  \item Resample the training data to create an equal balance of ``easy'' and ``hard'' training patterns. This step will be discussed further in \ref{subsec:resampling}.
  \item Initialize the grader, fit to the relabeled training data.
\end{enumerate}

\subsection{Evaluating New Inputs}
New inputs are evaluated as follows:

\begin{enumerate}
    \item Evaluate the input using the grader.
    \item Consider the output of the grader,
    \begin{enumerate}
        \item If ``easy'', re-evaluate with the base classifier and output the result.
        \item If ``hard'', re-evaluate with the deferral classifier and output the result.
    \end{enumerate}
\end{enumerate}

\subsection{Data Resampling} \label{subsec:resampling}
The percentage of training data relabeled as ``easy'' is equal to the training accuracy of the base classifier. For example, if the base classifier can successfully classify 95\% of training data, then 95\% of samples will be considered ``easy'' and the remaining 5\% will be considered ``hard''.

This means that the grader is often trained with a highly imbalanced dataset. In these cases, it is difficult for the grader to learn to recognize patterns in the minority class; it is not uncommon for the trained grader to be a trivial classifier which labels everything as ``easy''.

To counteract this, after relabeling we perform a resampling step to ensure an equal number of ``easy'' and ``hard'' training samples. In our experiments, we used the popular SMOTE (Synthetic Minority Over-sampling Technique) algorithm \cite{chawla2002smote}. SMOTE creates synthetic data by randomly selecting two samples belonging to the minority class and selecting a random point along the line connecting them. The effects of the resampling stage was examined within the context of the reject option in \cite{Vernon2022}.

\subsection{2-D Example}
\begin{figure*}[t]
\centering
\includegraphics[width=0.7\linewidth]{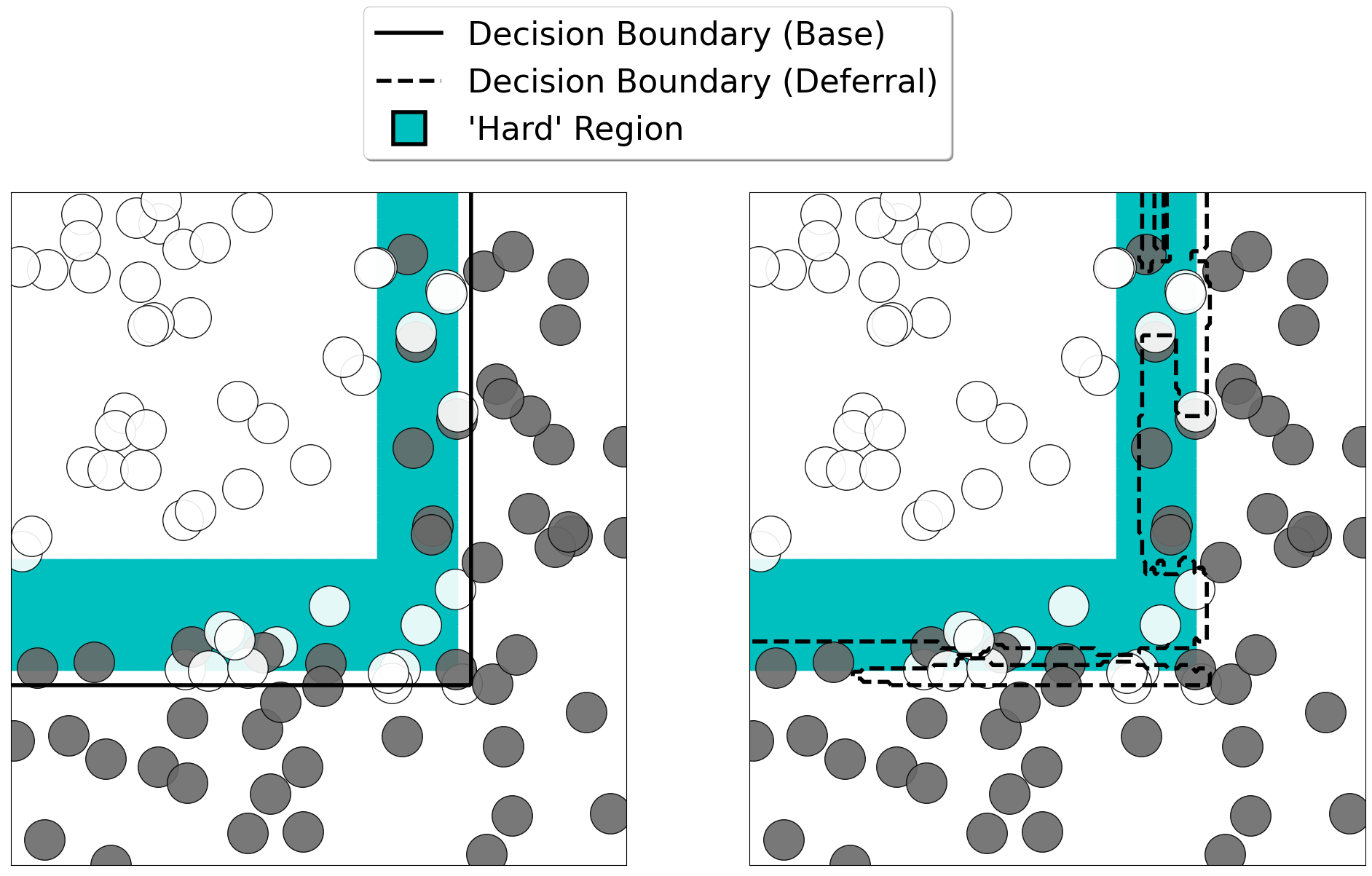}
\caption{Example decision boundaries when using a decision tree (left) and a random forest (right) classifier. Points which fall within the shaded region are considered ``hard'', and are labeled with the random forest. The remainder are considered ``easy'' and are evaluated using the decision tree. The shaded region itself is the output of another decision tree.}
\label{fig:demo}
\end{figure*}

\begin{figure*}[t]
\centering
\includegraphics[width=0.8\linewidth]{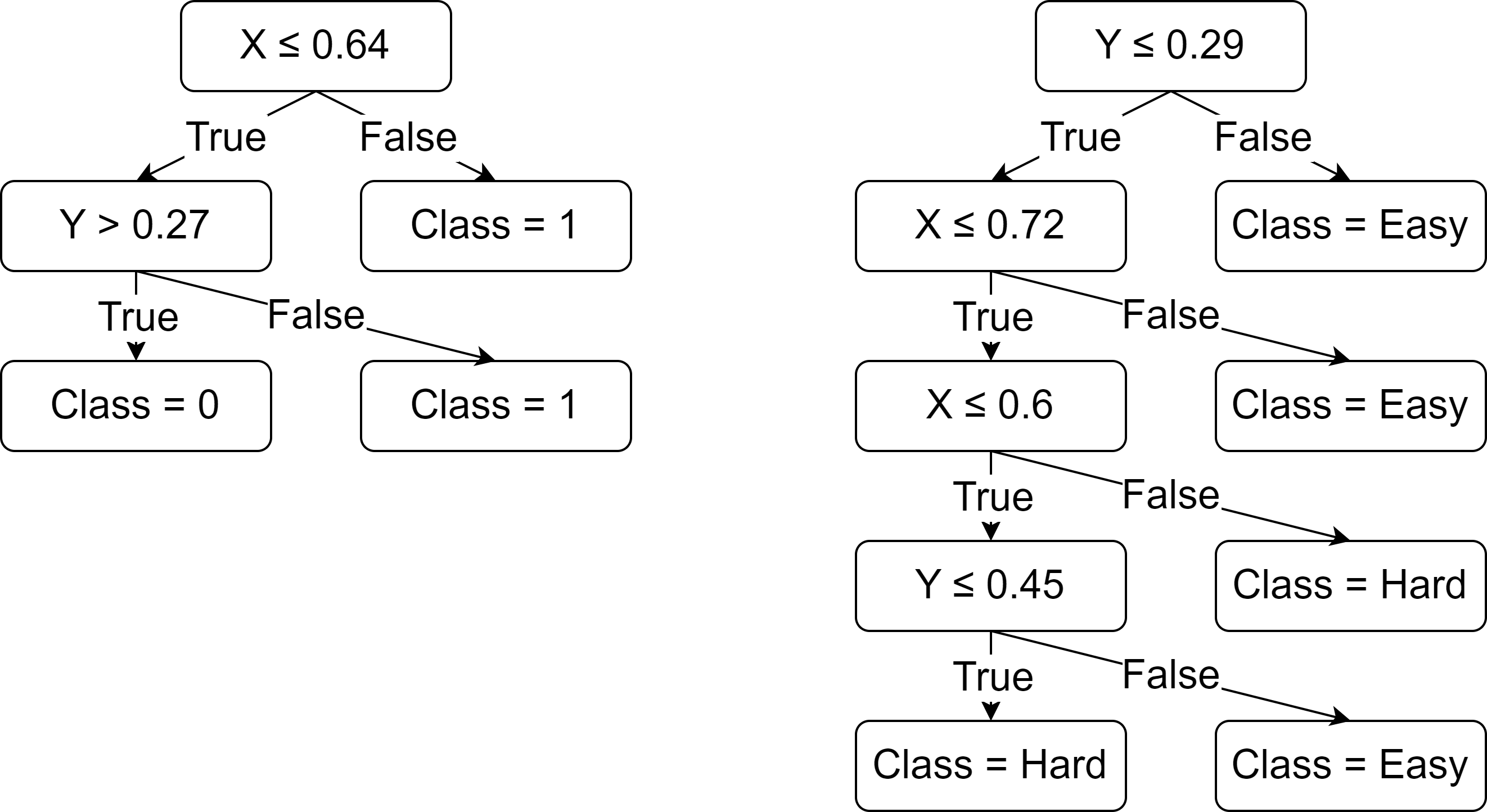}
\caption{Decision trees for the base classifier (left) and the grader (right). Patterns which the grader considers ``easy'' are evaluated using the base classifier, while ``hard'' patterns are evaluated using a more complex deferral classifier.}
\label{fig:demo_trees}
\end{figure*}

Figures \ref{fig:demo}-\ref{fig:demo_trees} demonstrate our method using a simple two-dimensional dataset.

The decision boundary of the base (decision tree) and deferral (random forest) classifiers are shown in Figure \ref{fig:demo} using solid and dashed lines respectively. The shaded region represents the area which the grader (decision tree) considers ``hard''. (i.e., The non-shaded region is considered ``easy''.) Figure \ref{fig:demo_trees} describes the decision trees of the base classifier and the grader. 

This example dataset consists of 100 points, 50 for each class. Independently, the base classifier can correctly classify 88\% of the points, and the deferral classifier can correctly classify 99\%. However, the base classifier has a much simpler decision boundary, and is more easily understood by humans.

The grader has identified the 23 points which fall within the shaded region as ``hard'', including every point which the base classifier mislabeled. When evaluating these points, the deferral classifier is used instead of the base classifier.

The final result is that 99\% of points are classified correctly. For 77\% of points, the interpretable base classifier is used. For the remaining 23\% of points, the black box deferral classifier is used, but the user can still easily understand the conditions when the deferral classifier is necessary.

\section{Computational Experiments}
In this section we show the benefit of our approach through computational experiments on real-world datasets.

\subsection{Experiment Design}
To conduct our experiments, we used Python 3 and the scikit-learn package, version 1.3 \cite{scikit-learn}. We used decision trees for the base classifier and the grader, with the \texttt{max\_depth} parameter set to 4 in both cases. We used a random forest classifier with default parameters (100 estimators, no maximum tree depth) for the deferral classifier.

We used 10 real-world based datasets from the OpenML repository \cite{OpenML2013}. The datasets are described in Table \ref{tab:datasets}. For simplicity, we selected datasets consisting of only numerical features, and with no missing values. Each dataset was tested using 10-fold cross validation, repeated 5 times, for a total of 50 runs.

We used the imbalanced-learn package, version 0.11, for the implementation of the SMOTE algorithm \cite{imbalanced-learn}.

\begin{table}[!t]
\centering
\vspace{5pt}
    \caption{Datasets used in computational experiments.}
    \label{tab:datasets}
\begin{tabular}{cc@{\hspace{2mm}}c@{\hspace{2mm}}c@{\hspace{2mm}}c@{\hspace{2mm}}c}
\hline
Dataset & Abbr. & \# Features & \# Patterns  & \# Classes  \\ \hline
Banknote Authentication & Bnk & 4 & 1372 & 2 \\
Blood Transfusion Service Center & Bld & 4 & 748 & 2 \\
Breast Cancer Wisconson (Diagnosis) & Brst & 30 & 569 & 2 \\
Climate Model Simulation Crashes & Clim & 20 & 540 & 2 \\
EEG Eye State & EEG & 14 & 14980 & 2 \\
Gas Sensor Array Drift & Gas & 128 & 13910 & 6 \\
Ionosphere & Ins & 34 & 351 & 2 \\
Landsat Satellite & Land & 36 & 6430 & 6 \\
Ozone Level Detection & Ozn & 72 & 2534 & 2 \\
QSAR Biodegradation & QSAR & 41 & 1055 & 2 \\
Spambase & Spm & 57 & 4601 & 2 \\
Steel Plates Faults & Stl & 27 & 1941 & 7 \\
Vehicle & Veh & 18 & 846 & 4 \\
Yeast & Yst & 8 & 1484 & 10 \\
\hline
\end{tabular}
%
\vspace{5mm}
\centering
\vspace{5pt}
    \caption{Classification performance in computational experiments}
    \label{tab:results}
\begin{tabular}{c@{\hspace{2mm}}c@{\hspace{2mm}}c@{\hspace{10mm}}c@{\hspace{2mm}}c@{\hspace{10mm}}c@{\hspace{2mm}}c}
\hline
Dataset & \multicolumn{2}{l}{Base Accuracy [\%]} & \multicolumn{2}{l}{Final Accuracy [\%]} & \multicolumn{2}{l}{Deferral Rate [\%]} \\ \cline{2-7} 
\multicolumn{1}{c}{} & Training & Test & Training & Test & Training & Test \\
\hline
Bnk & 96.50 & 95.42 & 99.79 & 98.57 & 21.82 & 21.78 \\
Bld & 80.51 & 77.62 & 90.05 & 75.26 & 45.42 & 45.36 \\
Brst & 98.47 & 93.28 & 99.98 & 93.92 & 8.89 & 9.13 \\
Clim & 94.57 & 90.22 & 99.66 & 90.48 & 19.64 & 20.78 \\
EEG & 70.67 & 70.19 & 93.51 & 87.92 & 62.32 & 62.57 \\
Gas & 73.27 & 72.96 & 96.16 & 95.50 & 37.38 & 37.44 \\
Ins & 94.57 & 87.35 & 99.62 & 89.35 & 22.79 & 26.56 \\
Lnd & 79.88 & 78.83 & 96.98 & 90.46 & 40.67 & 40.81 \\
Ozn & 95.02 & 92.97 & 99.40 & 93.99 & 25.34 & 26.39 \\
QSAR & 86.42 & 80.92 & 96.83 & 85.01 & 41.29 & 42.19 \\
Spm & 90.86 & 89.53 & 97.45 & 94.38 & 31.69 & 32.46 \\
Stl & 62.29 & 60.76 & 95.08 & 76.86 & 58.62 & 59.23 \\
Vhcl & 73.67 & 68.21 & 96.26 & 72.88 & 46.65 & 47.89 \\
Yst & 59.84 & 56.85 & 92.95 & 61.16 & 67.93 & 67.90 \\ 
\hline
\end{tabular}
\end{table}

\subsection{Experimental Results}
The experimental results are summarized in Table \ref{tab:results}. The values refer to the arithmetic means 
over all 50 runs for each dataset.

In Table \ref{tab:results}, ``Base Accuracy'' refers to the simple accuracy (i.e., percentage of correct predictions) of the base classifier against the entire training or testing set, independent of the outputs of the deferral classifier or grader. %

``Final Accuracy'' is to the simple accuracy of the ensemble, first evaluating patterns with the grader and then with either the base or deferral classifier as appropriate (see: Figure \ref{fig:evaluation}). The `Deferral Rate' is the percentage of patterns which were evaluated by the random forest classifier instead of the decision tree, i.e. were considered `hard' by the grader.

\begin{figure*}
\centering
\includegraphics[width=\linewidth]{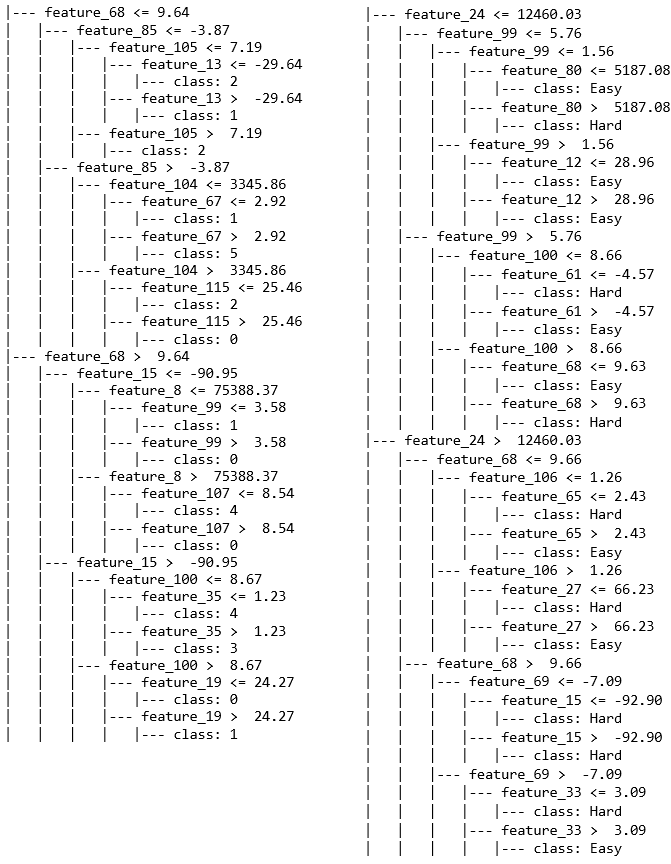}
\caption{A textual representation base classifier (left) and grader (right) decision trees for the ``Gas Sensor Array Drift'' dataset. While the dataset has 128 features, the majority of patterns can be correctly classified using only a small subset of features. Moreover, the set of patterns which are difficult to classify is just as easily described.}
\label{fig:gas_trees}
\end{figure*}

As an example, consider the ``Gas Sensor Array Drift'' dataset: 96.16\% of the training set and 95.50\% of testing set was classified correctly. The decision tree base classifier was used to evaluate roughly 62\% of patterns, with 37.38\% (train) and 37.44\% (test) deferred to the random forest classifier. For comparison, we can see that the decision tree on its own would have only classified 73.27\% (train) and 72.96\% (test) of patterns correctly. Samples of the discovered decision trees are shown in Figure \ref{fig:gas_trees}.

\section{Conclusion and Future Work}
In this paper, we introduced a classification method which combines both white and black box algorithms to improve accuracy while maintaining interpretability. Our method classifies patterns using either a highly-interpretable base classifier, or a highly-accurate deferral classifier. The determination for which classifier to use itself is made by a highly-interpretable grader, which judges inputs as either ``easy'' or ``hard''. This means the user, in addition to the classification result, receives either an explanation for the classification result, or an explanation for why a more complicated classifier needed to be used.

In our experiments, we used shallow decision trees for both white box models, and random forests for the black box model. In the future, we plan to experiment with many different types of classification algorithms, such as rule-based classifiers for the white box(es) and gradient boosted trees for the black box.

Additionally, we hope to develop a simple user interface for our design so that the user can quickly and easily visualize the structure of the white box model, the decision spaces of all three models, and experiment with different parameters (e.g., algorithm used for each step, maximum tree depth).

\section*{Acknowledgment}

This work was supported by JST SPRING, Grant Number JPMJSP2139, and the Japan Society for the Promotion of Science (JSPS) KAKENHI under Grant JP19K12159.

\small{
\bibliographystyle{splncs03_unsrt}
\bibliography{bib}

\begin{thebibliography}{10}
\providecommand{\url}[1]{\texttt{#1}}
\providecommand{\urlprefix}{URL }

\bibitem{Bohanec1994}
Bohanec, M., Bratko, I.: Trading accuracy for simplicity in decision trees.
  Machine Learning  15(3),  223 ^^e2^^80^^93 250 (1994)

\bibitem{Breiman2001}
Breiman, L.: Random forests. Machine Learning  45,  5--32 (2001)

\bibitem{Friedman2001}
Friedman, J.H.: Greedy function approximation: A gradient boosting machine.
  Annals of Statistics pp. 1189--1232 (2001)

\bibitem{Bentejac2021}
Bent{\'e}jac, C., Cs{\"o}rg{\H{o}}, A., Mart{\'\i}nez-Mu{\~n}oz, G.: A
  comparative analysis of gradient boosting algorithms. Artificial Intelligence
  Review  54,  1937--1967 (2021)

\bibitem{Silver2018}
Silver, D., Hubert, T., Schrittwieser, J., Antonoglou, I., Lai, M., Guez, A.,
  Lanctot, M., Sifre, L., Kumaran, D., Graepel, T., et~al.: A general
  reinforcement learning algorithm that masters chess, shogi, and go through
  self-play. Science  362(6419),  1140--1144 (2018)

\bibitem{Brown2020}
Brown, T., Mann, B., Ryder, N., Subbiah, M., Kaplan, J.D., Dhariwal, P.,
  Neelakantan, A., Shyam, P., Sastry, G., Askell, A., Agarwal, S.,
  Herbert-Voss, A., Krueger, G., Henighan, T., Child, R., Ramesh, A., Ziegler,
  D., Wu, J., Winter, C., Hesse, C., Chen, M., Sigler, E., Litwin, M., Gray,
  S., Chess, B., Clark, J., Berner, C., McCandlish, S., Radford, A., Sutskever,
  I., Amodei, D.: Language models are few-shot learners. In: Larochelle, H.,
  Ranzato, M., Hadsell, R., Balcan, M., Lin, H. (eds.) Advances in Neural
  Information Processing Systems. vol.~33, pp. 1877--1901. Curran Associates,
  Inc. (2020)

\bibitem{Lin2019}
Lin, T.C.: Artificial intelligence, finance, and the law. Fordham Law Review
  88,  531--551 (2019)

\bibitem{Longoni2019}
Longoni, C., Bonezzi, A., Morewedge, C.K.: Resistance to medical artificial
  intelligence. Journal of Consumer Research  46(4),  629--650 (2019)

\bibitem{Shin2021}
Shin, D.: The effects of explainability and causability on perception, trust,
  and acceptance: Implications for explainable {AI}. International Journal of
  Human-Computer Studies  146,  102551 (2021)

\bibitem{Wocjciech2021}
Samek, W., Montavon, G., Lapuschkin, S., Anders, C.J., M^^c3^^bcller, K.R.:
  Explaining deep neural networks and beyond: A review of methods and
  applications. Proceedings of the IEEE  109(3),  247--278 (2021)

\bibitem{Hendrickx2021}
Hendrickx, K., Perini, L., Van~der Plas, D., Meert, W., Davis, J.: Machine
  learning with a reject option: A survey. arXiv preprint arXiv:2107.11277
  (2021)

\bibitem{Chow1970}
Chow, C.: On optimum recognition error and reject tradeoff. IEEE Transactions
  on Information Theory  16(1),  41--46 (1970)

\bibitem{Ishibuchi1998}
Ishibuchi, H., Nakshima, T.: Fuzzy classification with reject options by fuzzy
  if-then rules. In: 1998 IEEE International Conference on Fuzzy Systems
  Proceedings. IEEE World Congress on Computational Intelligence (Cat. No.
  98CH36228). vol.~2, pp. 1452--1457. IEEE (1998)

\bibitem{Nojima2016}
Nojima, Y., Ishibuchi, H.: Multiobjective fuzzy genetics-based machine learning
  with a reject option. In: 2016 IEEE International Conference on Fuzzy Systems
  (FUZZ-IEEE). pp. 1405--1412. IEEE (2016)

\bibitem{Nojima2023}
Nojima, Y., Kawano, K., Shimahara, H., Vernon, E., Masuyama, N., Ishibuchi, H.:
  Fuzzy classifiers with a two-stage reject option. In: 2023 IEEE International
  Conference on Fuzzy Systems (FUZZ-IEEE). pp. 1--6. IEEE (2023)

\bibitem{chawla2002smote}
Chawla, N.V., Bowyer, K.W., Hall, L.O., Kegelmeyer, W.P.: {SMOTE}: Synthetic
  minority over-sampling technique. Journal of Artificial intelligence Research
   16,  321--357 (2002)

\bibitem{Vernon2022}
Vernon, E.M., Masuyama, N., Nojima, Y.: Error-reject tradeoff analysis on
  two-stage classifier design with a reject option. In: 2022 World Automation
  Congress (WAC). pp. 312--317. IEEE (2022)

\bibitem{scikit-learn}
Pedregosa, F., Varoquaux, G., Gramfort, A., Michel, V., Thirion, B., Grisel,
  O., Blondel, M., Prettenhofer, P., Weiss, R., Dubourg, V., Vanderplas, J.,
  Passos, A., Cournapeau, D., Brucher, M., Perrot, M., Duchesnay, E.:
  Scikit-learn: Machine learning in {P}ython. Journal of Machine Learning
  Research  12,  2825--2830 (2011)

\bibitem{OpenML2013}
Vanschoren, J., van Rijn, J.N., Bischl, B., Torgo, L.: Open{ML}: {N}etworked
  science in machine learning. SIGKDD Explorations  15(2),  49--60 (2013),
  \url{http://doi.acm.org/10.1145/2641190.264119}

\bibitem{imbalanced-learn}
Lema{{\^i}}tre, G., Nogueira, F., Aridas, C.K.: Imbalanced-learn: A python
  toolbox to tackle the curse of imbalanced datasets in machine learning.
  Journal of Machine Learning Research  18(17),  1--5 (2017),
  \url{http://jmlr.org/papers/v18/16-365.html}

\end{thebibliography}
}

\end{document}